\documentclass{article} 
\usepackage{iclr2027_conference,times}

\usepackage{amsmath,amsfonts,bm}

\def\eqref#1{equation~\ref{#1}}

\def\1{\bm{1}}

\DeclareMathAlphabet{\mathsfit}{\encodingdefault}{\sfdefault}{m}{sl}
\SetMathAlphabet{\mathsfit}{bold}{\encodingdefault}{\sfdefault}{bx}{n}

\usepackage{listings}
\usepackage{hyperref}
\usepackage{url}
\usepackage{xspace}
\usepackage{amsmath}
\usepackage{amssymb}
\usepackage{pifont}
\usepackage{booktabs}
\usepackage{xcolor}
\usepackage{colortbl}
\usepackage{graphicx}
\usepackage{algorithm}
\usepackage{wrapfig}
\usepackage{algpseudocode}
\usepackage{graphicx}
\usepackage{longtable,array}
\usepackage{enumitem}
\usepackage{subcaption}
\setlist[itemize]{noitemsep,leftmargin=*,topsep=0pt}
\setlist[enumerate]{noitemsep,leftmargin=*,topsep=0pt}

\usepackage{needspace}
\AtBeginDocument{%
}

\definecolor{heatabove}{HTML}{C0392B}
\definecolor{heatbelow}{HTML}{2471A3}
\newcommand{\heatspan}{1.5}   
\newcommand{\heatpeak}{60}    
\newcommand{\heatgamma}{0.75} 
\newcommand{\heat}[1]{%
  \edef\heattmp{\noexpand\cellcolor{%
    \ifdim #1pt<1pt heatbelow\else heatabove\fi%
    !\fpeval{round((min(1,abs(#1-1)/\heatspan)**\heatgamma)*\heatpeak,1)}}}%
  \heattmp #1}

\renewcommand{\checkmark}{\text{\ding{51}}}
\newcommand{\crossmark}{\text{\ding{55}}}

\newcommand{\ours}{\textsc{BabelArena}\xspace}

\newcommand{\oursmethod}{\textsc{BabelFlow}\xspace}

\newcommand{\opusfoureight}{Claude-Opus-4.8\xspace}
\newcommand{\qwenplusthreeseven}{Qwen-3.7-Plus\xspace}
\newcommand{\qwenmaxthreeseven}{Qwen-3.7-Max\xspace}
\newcommand{\qwenmaxthreeeight}{Qwen-3.8-Max\xspace}
\newcommand{\gptsolfivesix}{GPT-5.6-Sol\xspace}
\newcommand{\gptterrafivesix}{GPT-5.6-Terra\xspace}

\newcommand{\geminiflashthreefive}{Gemini-3.5-Flash\xspace}
\newcommand{\geminiflashthreeseven}{Gemini-3.7-Flash\xspace}
\newcommand{\geminiprothreeone}{Gemini-3.1-Pro\xspace}

\newcommand{\vitabench}{\textsc{VitaBench}\xspace}
\newcommand{\deepplanning}{\textsc{DeepPlanning}\xspace}
\newcommand{\taubench}{\textsc{$\tau^3$-Bench}\xspace}
\newcommand{\claweval}{\textsc{ClawEval}\xspace}

\newcommand{\passk}[1]{Pass\textasciicircum{}#1\xspace}
\newcommand{\passone}{\passk{1}}
\newcommand{\passthree}{\passk{3}}

\title{\ours: A Large-Scale Multilingual Benchmark for LLM Agents}

\author{\normalfont
\textbf{Peng Kuang}\textsuperscript{1,2,}\thanks{Equal contribution.}\textsuperscript{\phantom{*},}\thanks{Work done during an internship at Alibaba Token Hub, Alibaba Group.} \quad
\textbf{Yuchun Fan}\textsuperscript{1,3,}\footnotemark[1]\textsuperscript{\phantom{*},}\footnotemark[2] \quad
\textbf{Jiangnan Li}\textsuperscript{1,4,}\footnotemark[2] \quad
\textbf{Minghao Wu}\textsuperscript{1,\ding{41}} \\
\textbf{Jialong Tang}\textsuperscript{1} \quad
\textbf{Haoran Wei}\textsuperscript{1} \quad
\textbf{Weixuan Wang}\textsuperscript{5} \quad
\textbf{Jianhong Tu}\textsuperscript{1} \\
\textbf{Baosong Yang}\textsuperscript{1} \quad
\textbf{Tong Xiao}\textsuperscript{3} \\[1ex]
\textsuperscript{1}Alibaba Token Hub, Alibaba Group \quad \textsuperscript{2}University of Illinois Urbana-Champaign \\
\textsuperscript{3}Northeastern University \quad \textsuperscript{4}Monash University \quad \textsuperscript{5}Ant International
}

\iclrfinalcopy 
\begin{document}

\maketitle
\ificlrfinal
\begingroup
\renewcommand{\thefootnote}{\ding{41}}
\footnotetext{Corresponding author: \texttt{minghao.wu@alibaba-inc.com}}.
\endgroup
\fi

\begin{abstract}
Large language model (LLM) agents increasingly execute multi-step workflows through tool use and interaction with users and environments.
However, current agent evaluations are largely English-centric, limiting our understanding of agent capabilities in multilingual settings.
We introduce \oursmethod, a benchmark-general agentic workflow that adapts existing agent benchmarks to new languages by analyzing runtime dependencies, coordinating structure-preserving translation, and combining multi-layer verification with human review to preserve task and evaluation semantics.
Using \oursmethod, we construct \ours, a large-scale multilingual benchmark comprising 16,077 instances derived from 699 canonical tasks across four benchmark families, 13 domains, and 23 languages.
Experiments with five frontier models show that no single model dominates across benchmark families and that cross-language disparities extend well beyond task success.
Lower-resource languages exhibit distinct failure patterns, with larger shares of tool-use and control-flow errors rather than answer-quality errors alone, pointing to gaps in reliable task execution across the resource levels of these languages.
On the same tasks, agents in low-resource languages also consume substantially more tokens than in English (up to roughly twice the input) without proportional increases in interaction length, and language consistency degrades further on tasks requiring structured output, where switches are directed overwhelmingly toward English.
We believe \ours provides a foundation for advancing research on reliable and efficient multilingual agents.\footnote{The project is available at: \url{https://github.com/QwenLM/BabelArena}}
\end{abstract}

\section{Introduction}

Large language models (LLMs) are increasingly used as agents that go beyond answering questions to execute multi-step workflows through tool use and interaction with users and environments~\citep{kuang2026kvprmefficientprocessreward}.
To serve users in different languages, agents need to do more than produce fluent responses~\citep{liu2024agentbench,DBLP:conf/iclr/ZhouX0ZLSCOBF0N24,patil2025bfcl}.
They must interpret requests correctly, act on tool feedback, and maintain task constraints throughout the workflow~\citep{luo2026lostexecution,nguyen2026seataubench}.
Therefore, evaluation must assess whether agents can complete these workflows reliably across languages, rather than judging response fluency alone.

However, existing agent benchmarks provide limited support for such multilingual end-to-end evaluation, as most remain concentrated in English or Chinese~\citep{hofman2026maps,kim2026gaiav2lilt,caciolai2026omnilingualgaia2}.
Recent multilingual benchmarks attempt to address this gap, but most adapt only selected components of the source benchmarks and cover a limited range of languages and domains~\citep{hofman2026maps,kim2026gaiav2lilt,caciolai2026omnilingualgaia2,luo2026lostexecution,nguyen2026seataubench,li2026polyworkbench,jin2026breakingbabelselfevolvingmultiagent} (\autoref{tab:multilingual-benchmark-comparison}).
Extending this coverage is particularly challenging because agent benchmarks are not self-contained input--output datasets, but executable environments composed of multiple interacting components~\citep{DBLP:conf/iclr/ZhouX0ZLSCOBF0N24,yao2024taubench,patil2025bfcl}.
Across these components, task instructions, tools, databases, simulated users, and evaluators are connected through shared entities and constraints that must remain consistent.
Translating these components independently can break cross-component consistency, causing execution or scoring failures even when the agent behaves correctly~\citep{kim2026gaiav2lilt,luo2026lostexecution,nguyen2026seataubench}.
Reliable cross-language comparison therefore requires coordinated adaptation that preserves these dependencies while keeping task and evaluation semantics aligned across languages.

To address these challenges, we introduce \oursmethod, a benchmark-general agentic workflow for adapting existing agent benchmarks to new languages while preserving task and evaluation semantics.
A coding agent first analyzes runtime dependencies to identify language-bearing fields and distinguish translatable content from execution-critical values.
Guided by these dependencies, a translator agent translates fields consistently across components and reuses canonical translations for shared entities, while the workflow reconstructs the translated fields in their original structures.
Independent verifier agents then check structural integrity, cross-component consistency, semantic fidelity, and behavioral equivalence, with human reviewers adjudicating their findings and guiding revisions before a language variant of the benchmark is accepted.

Applying \oursmethod to \vitabench~\citep{he2025vitabench}, \deepplanning~\citep{zhang2026deepplanning}, \taubench~\citep{yao2024taubench,DBLP:journals/corr/abs-2506-07982,shi2026tauknowledge}, and \claweval~\citep{ye2026claweval}, we construct \ours, a multilingual benchmark covering interactive services, long-horizon planning, knowledge-grounded tool use, and heterogeneous autonomous tasks.
With 699 distinct tasks across 13 domains and 23 languages, \ours offers more tasks than most prior benchmarks, the broadest language and domain coverage, and the largest total of 16,146 task instances among the benchmarks compared in \autoref{tab:multilingual-benchmark-comparison}.
Task identities and source evaluation criteria remain fixed across languages, enabling controlled comparisons of agent performance.

\begin{table}[t]
    \centering
    \setlength{\tabcolsep}{6.5pt}
    \caption{Comparison of multilingual agent benchmarks. \checkmark{}/\crossmark{} denote adapted/unadapted components: system, simulator, and evaluation prompts (\emph{Prompt}), task inputs (\emph{Task}), tool descriptions (\emph{Tool}), and environment content (\emph{Env.}). \emph{Tasks} counts distinct tasks; \emph{Dom.}, domains; \emph{Lang.}, languages; and \emph{Inst.}, total instances across languages. Boldface marks \ours.}
    \label{tab:multilingual-benchmark-comparison}
    \begin{tabular}{@{}lcccccccc@{}}
        \toprule
        & \multicolumn{4}{c}{Adapted Components} & \multicolumn{4}{c}{Statistics} \\
        \cmidrule(lr){2-5}\cmidrule(l){6-9}
        Benchmark & Prompt & Task & Tool & Env. & Tasks & Dom. & Lang. & Inst. \\
        \midrule
        MAPS~\citeyearpar{hofman2026maps}          & \crossmark & \checkmark & \crossmark & \crossmark & 805 & \phantom{0}4 & 12 & \phantom{0}9,660 \\
        GAIA-v2-LILT~\citeyearpar{kim2026gaiav2lilt}  & \checkmark & \checkmark & \crossmark & \crossmark & 165 & \phantom{0}1 & \phantom{0}6 & \phantom{00,}990 \\
        OmnilingualGAIA2~\citeyearpar{caciolai2026omnilingualgaia2} & \checkmark & \checkmark & \crossmark & \checkmark & 640 & \phantom{0}1 & 11 & \phantom{0}7,040 \\
        MLCL~\citeyearpar{luo2026lostexecution}          & \crossmark & \checkmark & \crossmark & \crossmark & 200 & \phantom{0}1 & \phantom{0}4 & \phantom{00,}800 \\
       SEATauBench~\citeyearpar{nguyen2026seataubench}   & \checkmark & \checkmark & \checkmark & \checkmark & 278 & \phantom{0}3 & \phantom{0}6 & \phantom{0}1,668 \\
        PolyWorkBench~\citeyearpar{li2026polyworkbench} & \checkmark & \checkmark & \crossmark & \checkmark & \phantom{0}67 & \phantom{0}5 & 10 & \phantom{000}67 \\
        \midrule
        \textbf{\ours} & \checkmark & \checkmark & \checkmark & \checkmark & \textbf{699} & \textbf{13} & \textbf{23} & \textbf{16,077} \\
        \bottomrule
    \end{tabular}

\end{table}

We first validate \oursmethod through a controlled ablation on 40 tasks adapted into Chinese, finding fewer broken references and behavioral mismatches than with independent field translation (\autoref{sec:experiments_method}).
We then evaluate five frontier models on \ours and find that no single model leads across all four benchmark families, while all five models have lower average task success and language consistency in the low-resource group than in the high-resource group (\autoref{sec:experiments_benchmark}).
Among sampled \vitabench failures, Thai and Tamil show larger shares of tool-use and control-flow errors than English and Chinese, suggesting that cross-language disparities extend beyond answer quality to execution reliability (\autoref{sec:analysis}).
On the same \vitabench tasks, low-resource trajectories consume $1.69$--$2.06\times$ as many total input tokens as their English counterparts with similar numbers of turns, while shopping tasks in \deepplanning also incur more turns and tool calls.
English accounts for 91.2\% of annotated switches among sampled language-inconsistent \deepplanning and \claweval trajectories, with inconsistency concentrated in assistant responses and textual prefixes before tool calls, respectively.

Our contributions are threefold:
\begin{itemize}
    \item We introduce and validate \oursmethod, a benchmark-general agentic workflow for adapting agent benchmarks across languages while preserving the semantics (\autoref{sec:ada_pipeline} and \autoref{sec:experiments_method}).
    \item We construct \ours, a multilingual agent benchmark with aligned tasks and evaluation criteria across languages (\autoref{sec:our_benchmark}).
    \item We evaluate five frontier models, revealing language-dependent differences in task success, language consistency, failure patterns, and execution costs (\autoref{sec:experiments_benchmark} and \autoref{sec:analysis}).
\end{itemize}

\section{Related Work}

\paragraph{Agent Evaluation}
Agent benchmarks increasingly measure closed-loop behavior in executable environments rather than isolated final answers \citep{liu2024agentbench,DBLP:conf/iclr/ZhouX0ZLSCOBF0N24,patil2025bfcl}.
Our four source benchmarks cover interactive services, constrained planning, knowledge-grounded tool use, and autonomous tasks~\citep{he2025vitabench,zhang2026deepplanning,yao2024taubench,DBLP:journals/corr/abs-2506-07982,shi2026tauknowledge,ye2026claweval}.

\paragraph{Multilingual Evaluation}
Conventional multilingual benchmarks use self-contained input--output pairs~\citep{bandarkar-etal-2024-belebele,kulkarni-etal-2025-massive}.
Agent benchmarks also expose language through prompts, tools, environments, and evaluators.
MAPS and MLCL mainly translate task inputs~\citep{hofman2026maps,luo2026lostexecution}, while GAIA-v2-LILT also adapts evaluation~\citep{kim2026gaiav2lilt}.
PolyWorkBench and OmnilingualGAIA2 additionally localize environments and evaluators, with the latter calibrating its multilingual verifier~\citep{li2026polyworkbench,caciolai2026omnilingualgaia2}.
SEATauBench adapts all components but covers fewer domains and languages than \ours~\citep{nguyen2026seataubench}.

\paragraph{Software Localization}
Software localization separates translatable resources from program logic~\citep{PirroneD24,xia2013softwareinternationalization,wang2013externalize}.
Its practice further treats validation as more than a fluency check: internationalization testing examines whether software works properly in a specific language and region~\citep{CoutoMG25,FelipeMC24}.

\paragraph{Ours}
\oursmethod adapts interdependent benchmark components while preserving execution and evaluation semantics.
Applied to four benchmark families, it produces \ours, with the broadest language and domain coverage and the most instances among benchmarks in \autoref{tab:multilingual-benchmark-comparison}.

\section{Adapting Agent Benchmarks From English to Multilingual}
\label{sec:ours_method}

We first outline the challenges of adapting agent benchmarks across languages (\autoref{sec:challenges}).
We then introduce \oursmethod, an agentic workflow that combines structured translation with multi-layer verification and human review (\autoref{sec:ada_pipeline}).
Finally, we apply \oursmethod to construct \ours (\autoref{sec:our_benchmark}).

\subsection{New Era, New Challenges}
\label{sec:challenges}

Multilingual chatbot benchmarks are commonly built by translating self-contained input--output pairs to reduce authoring costs.
However, agent benchmarks link execution-critical text across tasks, tools, databases, simulated users, and graders.
Adapting these executable environments introduces a few new challenges:

\begin{enumerate}[label=\textbf{(C\arabic*)}, ref=C\arabic*]
    \item \label{ch:exec} \textbf{Executability Preservation}: Like software localization, agent benchmark adaptation must preserve executability. Therefore, we must examine their code, data, and evaluators to determine what can be safely translated.
    \item \label{ch:consistency} \textbf{Translation Consistency}: Shared entities and values, such as product names and tool names, must remain consistent across databases, tool responses, and ground truth. Otherwise, execution may fail, or correct agent behavior may be scored as incorrect.
    \item \label{ch:validity} \textbf{Evaluation Validity}: Output parsers designed for English may fail when other languages use different word forms or word order. Translating instructions for LLM-simulated users can change their behavior, while LLM judges vary in accuracy and strictness across languages. 
    \item \label{ch:efficiency} \textbf{Evaluation Efficiency}: Multi-turn rollouts, tool execution, environment simulation, and model-based judging make each evaluation costly. Multiplying this cost across languages can make exhaustive evaluation infeasible.
\end{enumerate}

\ref{ch:exec}--\ref{ch:validity} motivate the adaptation framework (\autoref{sec:ours_method}) to preserve each language variant's fidelity, while \ref{ch:efficiency} guides benchmark composition (\autoref{sec:our_benchmark}) to control multilingual evaluation cost.

\begin{figure}[t]
    \centering
    \includegraphics[width=0.95\linewidth]{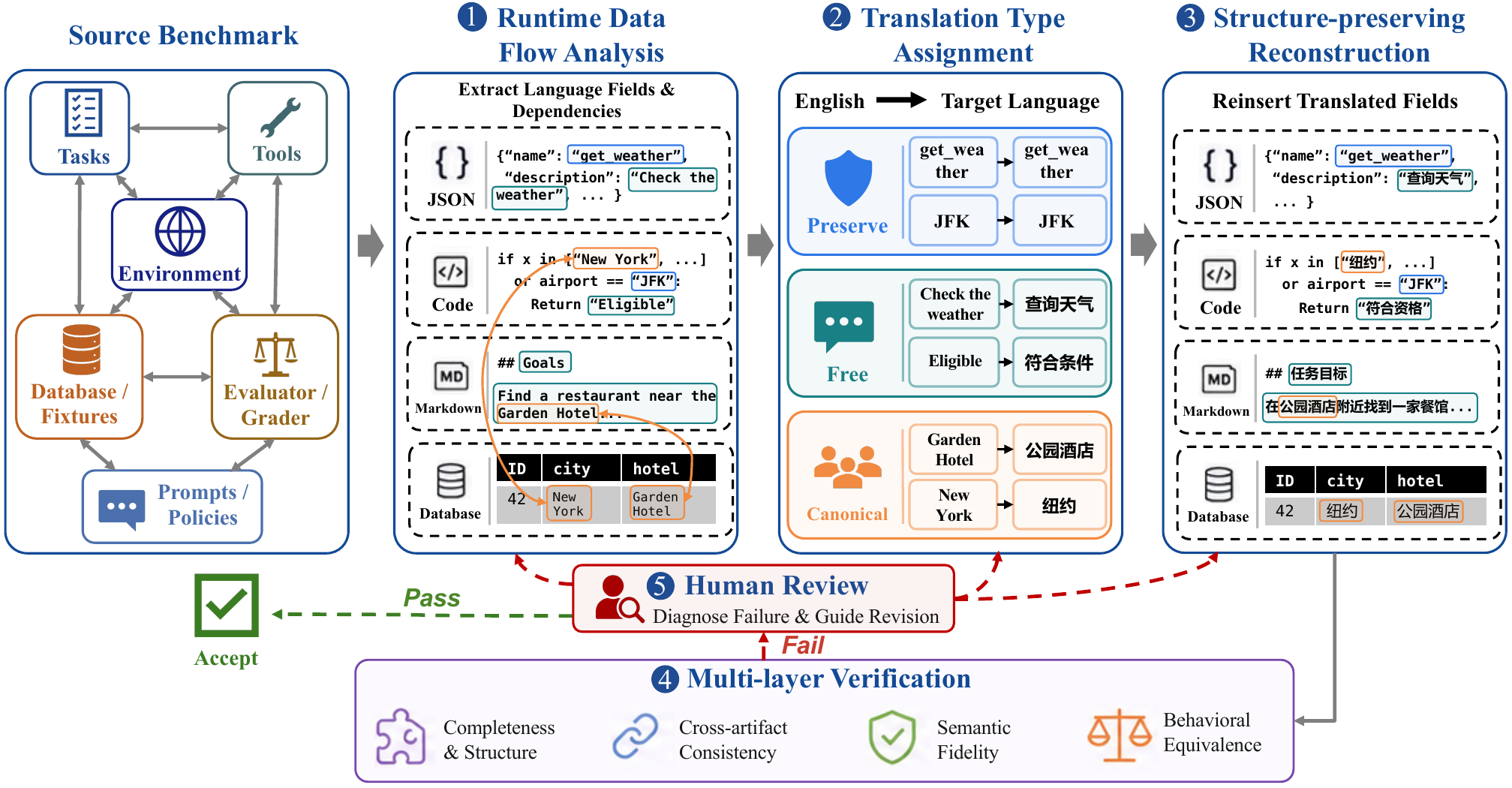}
    \caption{Overview of \oursmethod. Runtime data-flow analysis identifies text fields and dependencies for translation by type and reconstruction in their original structures. Multi-layer verification checks adaptation quality. Human review guides revisions to analysis and translation.}

    \label{fig:overview}
\end{figure}

\subsection{\oursmethod: Agentic Workflow Adapting Benchmarks Beyond English}
\label{sec:ada_pipeline}

Given a source benchmark $\mathcal{B}^{s}$ in language $\ell_{s}$ and a target language $\ell_{t}$, \oursmethod constructs an independently executable variant $\mathcal{B}^{t}$ in $\ell_{t}$.
As shown in \autoref{fig:overview}, the analyzer $\mathcal{A}$ (\opusfoureight) identifies text fields and cross-artifact dependencies and assigns translation types (Steps 1 and 2), while the translator $\mathcal{T}$ (\qwenplusthreeseven) translates the fields for reconstruction in their original structures (Step 3).
The verifier $\mathcal{V}$ comprises three heterogeneous agents (\gptsolfivesix, \qwenmaxthreeseven, and \opusfoureight) that independently check the variant across multiple layers and pool their findings (Step 4).
Human reviewers triage these findings and guide revisions by $\mathcal{A}$ and $\mathcal{T}$ (Step 5).
The variant $\mathcal{B}^{t}$ is accepted only after every layer passes and a reviewer signs off.
Steps 1--3 establish \ref{ch:exec} and \ref{ch:consistency} by construction, and Steps 4 and 5 re-test them together with \ref{ch:validity} by execution.

\paragraph{Step 1: Runtime data flow analysis}
The analyzer $\mathcal{A}$, a coding agent, reads and runs the source benchmark to trace its evaluation data flow and inventory every text field accessed at runtime, whether in data files or code: task specifications, policies and system prompts, simulated-user instructions, tool documentation and schemas, databases and fixtures, tool observations, output parsers, rubrics, judge prompts, and grader constants.
It then links fields that must agree across artifacts, such as product names shared by requests, database rows, tool responses, and ground truth, or category labels shared by task prose and evaluator lexical matchers.
These dependencies determine which fields can be safely translated.
The analysis yields a typed intermediate representation $\mathcal{I}^{s} = \mathcal{A}(\mathcal{B}^{s}) = \{e^{s}_{i}\}_{i=1}^{N}$, where each element $e^{s}_{i}$ represents a source-language text field and records its source string, artifact location, and the fields it must agree with.

\paragraph{Step 2: Translation type assignment}
The analyzer $\mathcal{A}$ assigns each element $e^{s}_{i} \in \mathcal{I}^{s}$ a translation type $t_{i}$ to guide the translator $\mathcal{T}$ when building the target variant:

\begin{itemize}
    \item \textbf{\textsc{Preserve}:} Keep byte-identical, with a declared reason that is either \emph{executable}, covering tool and argument names, schema keys, IDs, enum codes, dates, and evaluator control tokens, or \emph{source asset}, covering reference documents, images, audio, video, and other task-defining source media.
    \item \textbf{\textsc{Translate-Free}:} Translate in local context, covering instructions, policies, personas, descriptions, and rubric prose.
    \item \textbf{\textsc{Translate-Canonical}:} Translate once and reuse, covering cities, products, display labels, weekdays, and evaluator-matched terms.
\end{itemize}

The \textsc{Preserve} reasons distinguish execution-critical values, which must remain unchanged for the benchmark to run, from source assets whose translation would change the task.
Recording the reason makes exemptions auditable and distinguishes retained assets from incomplete translations.
Only source assets count against a variant's localization coverage.
These types guide $\mathcal{T}$ in Step 3: translate ordinary prose, reuse canonical translations across artifacts, and retain execution-critical values and declared source assets.

\paragraph{Step 3: Structure-preserving field extraction and reconstruction}
To protect artifact structure and out-of-scope content such as JSON keys and code, \oursmethod passes only extracted fields to the translator $\mathcal{T}$.
It copies the source artifact, extracts fields marked for translation, and uses placeholders to protect identifiers, templates, markup, and exact-format examples.
The translator renders free prose in $\ell_{t}$ and reuses shared-entity translations from the canonical translation map $\mathcal{C}$.
The results are written back to their original positions in the copied structure, preserving keys, nesting, and executable values.
Translating each element $e^{s}_{i}$ according to its type $t_{i}$ yields its target-language counterpart $e^{t}_{i}$ and the representation $\mathcal{I}^{t} = \mathcal{T}(\mathcal{I}^{s}, \ell_{t}) = \{e^{t}_{i}\}_{i=1}^{N}$.
Reconstruction following cross-artifact dependencies produces an adapted benchmark $\mathcal{B}^{t}$ that executes in $\ell_{t}$.

\paragraph{Step 4: Multi-layer verification}
Translation fluency alone does not establish benchmark validity. Three agents backed by \gptsolfivesix, \qwenmaxthreeseven, and \opusfoureight form the verifier $\mathcal{V}$: each independently executes $\mathcal{B}^{s}$ and $\mathcal{B}^{t}$ side by side and checks four layers:
\begin{itemize}
    \item \textbf{Completeness and structure.} Required artifacts exist, parse correctly, and retain source keys, types, nesting, and placeholders.
    \item \textbf{Cross-artifact consistency.} Protected values remain unchanged, shared entities use consistent translations, and references resolve across tasks, tools, databases, parsers, and graders.
    \item \textbf{Semantic fidelity.} Translations preserve intent, constraints, and information, with particular attention to fields affecting task difficulty or scoring.
    \item \textbf{Behavioral equivalence.} The verifier compares tool results, state changes, terminal states, parser outputs, and grader decisions under identical controlled actions in both environments.
\end{itemize}

The agents do not see one another's reports. Their findings are merged into a failure set $\mathcal{F} = \mathcal{V}(\mathcal{B}^{s}, \mathcal{B}^{t})$, retaining every issue flagged by any agent. Each entry records the offending field, verification layer, reasoning, and side-by-side evidence.
The first two layers re-check \ref{ch:exec} and \ref{ch:consistency} on the built variant, where mistyped fields may still parse and read fluently.
The last two target \ref{ch:validity}.

\paragraph{Step 5: Human review}
Reviewing $\mathcal{B}^{t}$ line by line is infeasible because its interdependent text fields span tasks, prompts, tools, databases, and evaluators, and many defects surface only at runtime.
The verifier group identifies candidate defects and provides reasoning.
Human reviewers confirm findings before revision because verifiers can misjudge.
Reviewers inspect every failure in $\mathcal{F}$, discard false alarms, and trace confirmed defects to their source stage to produce revision guidance $\mathcal{R}$: mistyped or missing elements return to the analyzer $\mathcal{A}$ (Steps 1 and 2), while fidelity or canonical-consistency errors return to the translator $\mathcal{T}$ with updates to the canonical translation map $\mathcal{C}$ (Step 3).
Analysis, translation, verification, and review repeat until $\mathcal{F} = \emptyset$, after which reviewers spot-check a random sample of tasks end to end and sign off on the variant.

\subsection{\ours: Agent Benchmark for Multilingual Agents}
\label{sec:our_benchmark}

\paragraph{Benchmark Overview} We instantiate \oursmethod on four source benchmarks with complementary interaction structures and target capabilities: \vitabench for versatile interactive tasks in food delivery, in-store consumption, online travel and cross-scenario; \deepplanning for long-horizon travel and shopping planning; the text component of \taubench for simulated-user interaction, tool calling, knowledge retrieval, and domain-policy compliance; and \claweval for service orchestration, multimodal perception and generation, and professional dialogue. To control the evaluation cost (\ref{ch:efficiency}), we randomly select 50\% tasks from each domain of the benchmark families.
The resulting \ours release covers 699 canonical task identities and 16,077 instances (\autoref{tab:multilingual-benchmark-comparison}), each benchmark spanning the same 23 languages (\autoref{tab:language-coverage}).
Our selected languages cover 15 scripts and high-, medium-, and low-resource levels, capturing variation in writing direction, word delimiters, and glyph shaping.

\begin{table}[t]
    \centering
    \small
    \setlength{\tabcolsep}{14pt}
    \caption{The 23 languages in \ours, grouped by LLM resource level. Codes combine the ISO 639-3 language tag with the ISO 15924 script tag.}
    \label{tab:language-coverage}
    \begin{tabular}{@{}cc@{\hspace{12pt}}cc@{\hspace{12pt}}cc@{}}
        \toprule
        \multicolumn{2}{c}{High resource (7)} & \multicolumn{2}{c}{Medium resource (8)} & \multicolumn{2}{c}{Low resource (8)} \\
        \cmidrule(r){1-2}\cmidrule(lr){3-4}\cmidrule(l){5-6}
        English  & \texttt{eng\_Latn} & Korean     & \texttt{kor\_Hang} & Belarusian      & \texttt{bel\_Cyrl} \\
        Chinese  & \texttt{zho\_Hans} & Indonesian & \texttt{ind\_Latn} & Eastern Panjabi & \texttt{pan\_Guru} \\
        Japanese & \texttt{jpn\_Jpan} & Hindi      & \texttt{hin\_Deva} & Kazakh          & \texttt{kaz\_Cyrl} \\
        French   & \texttt{fra\_Latn} & Thai       & \texttt{tha\_Thai} & Khmer           & \texttt{khm\_Khmr} \\
        Russian  & \texttt{rus\_Cyrl} & Hebrew     & \texttt{heb\_Hebr} & Burmese         & \texttt{mya\_Mymr} \\
        Arabic   & \texttt{arb\_Arab} & Turkish    & \texttt{tur\_Latn} & Lao             & \texttt{lao\_Laoo} \\
        Spanish  & \texttt{spa\_Latn} & Vietnamese & \texttt{vie\_Latn} & Tamil           & \texttt{tam\_Taml} \\
                 &                    & Malay      & \texttt{zsm\_Latn} & Telugu          & \texttt{tel\_Telu} \\
        \bottomrule
    \end{tabular}%
\end{table}

\paragraph{Evaluation}
We retain each source benchmark's scoring procedure, translating evaluation prompts as needed, and run three independent trials per task.
From task-completion rewards, we report \passone, the average single-trial success rate, and \passthree, the fraction of tasks solved in all three independent trials.
For each trajectory $\tau$, \geminiflashthreeseven assigns $\mathrm{LC}(\tau,\ell_{t})=1$ if the agent's \textit{user-visible text} consistently uses the target language $\ell_{t}$, and $0$ otherwise.
This check covers assistant responses, textual prefixes before tool calls, and natural-language content in deliverables (\autoref{app:language-consistency}).
We average LC across trials and report all three metrics separately.

\paragraph{Quality Assurance}
We assess translation quality by randomly sampling 200 translatable texts per language, yielding 4,600 records across the 23 languages.
Each translation is scored on a scale from 1 to 5 by human language experts for the seven high-resource languages and by \gptsolfivesix for the sixteen medium- and low-resource languages.
The mean scores are 4.28, 4.02, and 3.91 for high-, medium-, and low-resource languages, respectively.
Further details are provided in \autoref{app:annotation}.

\section{Experiments}
\label{sec:experiments_all}

We first validate \oursmethod by measuring adaptation defects and construction token costs (\autoref{sec:experiments_method}).
We then evaluate five frontier models on \ours, comparing task performance and language consistency across benchmark families and language resource groups (\autoref{sec:experiments_benchmark}).

\subsection{Validating \oursmethod}
\label{sec:experiments_method}

\begin{wraptable}{r}{0.4\textwidth}
    \centering
    \small
    \setlength{\tabcolsep}{2pt}
    \caption{Ablation of \oursmethod on 40 tasks adapted into Chinese. \emph{Br. ref.} and \emph{Be. mis.} report percent of tasks with broken references and behavioral mismatches; token costs are relative to independent translation. Lower is better.}
    \label{tab:babelflow-validation}
    \begin{tabular}{@{}lccc@{}}
        \toprule
        Method & Br. ref. & Be. mis. & Cost (tok.) \\
        \midrule
        Independent & 20.0 & 5.0 & 1.00 \\
        + Canonical & \phantom{0}7.5 & 2.5 & 1.78 \\
        \oursmethod & \phantom{0}2.5 & 0.0 & 3.35 \\
        \bottomrule
    \end{tabular}
\end{wraptable}

\paragraph{Setup}
We adapt 40 canonical tasks (10 per benchmark family) into Chinese, comparing \emph{independent field translation}, \emph{canonical translation} reusing $\mathcal{C}$, and \emph{\oursmethod}, which adds multi-layer verification and human-guided revision.
We report the percentage of tasks affected by each defect type: \emph{broken references} (inconsistent entities or unresolved references) and \emph{behavioral mismatches} (inconsistent tool, state, parser, or grading outcomes under semantically equivalent controlled actions across languages).

\paragraph{Results}
\autoref{tab:babelflow-validation} highlights the effectiveness of \oursmethod: compared with independent translation, it lowers the broken reference rate from 20.0\% (eight failed tasks) to 2.5\% (one failed task), with no behavioral mismatches observed among the 40 tasks.
Canonical translation alone leaves higher defect rates (7.5\% for broken references and 2.5\% for behavioral mismatches), supporting the value of verification and human-guided revision.
These gains come at a higher construction cost of $3.35\times$ the baseline token usage.
The remaining reference errors highlight the challenge of preserving consistency across benchmark components during multilingual adaptation.

\subsection{Evaluating Multilingual Agents}
\label{sec:experiments_benchmark}

\paragraph{Setup}
We evaluate five frontier models on \ours: \opusfoureight, \gptterrafivesix, \geminiprothreeone, \qwenmaxthreeseven, and \qwenmaxthreeeight.
All five use medium reasoning effort to control costs.
We use \geminiflashthreefive as the simulated user agent on all four benchmarks and as the convert agent for \deepplanning, mapping free-form plans to the required structured format.
\gptsolfivesix serves as the LLM judge for \vitabench, \taubench, and \claweval, while \deepplanning uses deterministic rules without an LLM judge.
All auxiliary agents run with thinking disabled, and all models use temperature 0.

\paragraph{Model Performance per Benchmark Family}
\autoref{tab:performance-benchmark-family} reports per-family results on \ours.
\qwenmaxthreeeight generally achieves the best overall performance across the benchmark families, leading on accuracy for both reasoning-heavy tasks, \vitabench and \deepplanning, but no single model wins on all the benchmarks, suggesting the diversity of our benchmark.
However, \qwenmaxthreeeight leads on accuracy, yet it falls behind on language consistency. 
We also observe that language consistency on \deepplanning is significantly worse than on the other benchmarks, and we dive into this in \autoref{sec:analysis}.

\begin{table}[t]
    \centering
    \small
    \caption{Model performance by benchmark family on \ours. We report \passone, \passthree, and language consistency (LC) in percent. Higher is better. Best results are in \textbf{bold}.}
    \label{tab:performance-benchmark-family}
    \setlength{\tabcolsep}{3pt}
    \begin{tabular}{@{}l*{12}{c}@{}}
        \toprule
        & \multicolumn{3}{c}{\vitabench} & \multicolumn{3}{c}{\deepplanning} & \multicolumn{3}{c}{\taubench} & \multicolumn{3}{c}{\claweval} \\
        \cmidrule(lr){2-4}\cmidrule(lr){5-7}\cmidrule(lr){8-10}\cmidrule(l){11-13}
        Model & \passone & \passthree & LC & \passone & \passthree & LC & \passone & \passthree & LC & \passone & \passthree & LC \\
        \midrule
        \opusfoureight     & 36.9 & 20.1 & 98.5 & 36.2 & 17.2 & 61.8 & 62.6 & 49.5 & \textbf{95.0} & \textbf{55.8} & \textbf{43.0} & 96.3 \\
        \gptterrafivesix   & 29.6 & 15.1 & 99.1 & 22.5 & 10.7 & \textbf{90.0} & 69.2 & 38.5 & 82.5 & 46.1 & 33.5 & 98.7 \\
        \geminiprothreeone & 34.3 & 17.7 & \textbf{99.9} & 18.6 & \phantom{0}5.6 & 74.3 & \textbf{71.4} & 46.9 & 86.2 & 41.5 & 29.3 & \textbf{98.8} \\
        \qwenmaxthreeseven & 36.1 & 18.9 & 98.6 & 41.6 & 23.8 & 57.5 & 59.0 & 49.1 & 92.8 & 32.4 & 24.6 & 98.5 \\
        \qwenmaxthreeeight & \textbf{42.3} & \textbf{25.0} & 94.7 & \textbf{46.0} & \textbf{26.5} & 75.8 & 67.6 & \textbf{57.0} & 87.6 & 40.8 & 27.1 & 88.7 \\
        \bottomrule
    \end{tabular}%
\end{table}

\paragraph{Model Performance per Language Group}
\autoref{tab:performance-language-group} shows that agent performance generally decreases as the language resource level declines.
From high- to low-resource languages, both \passone and \passthree decrease substantially across the five models.
Language consistency also declines for every model, with LC dropping by 3.7--14.8 points.
While \opusfoureight leads both task-completion metrics across all three groups, \gptterrafivesix achieves the highest LC, indicating that strong task performance does not necessarily imply consistent use of the target language.
These results reveal disparities across resource groups in both task completion and language adherence, motivating stronger support for agents operating in lower-resource languages.

\begin{table}[t]
    \centering
    \small
    \caption{Model performance by language resource group on \ours. We report \passone, \passthree, and language consistency (LC) in percent. Higher is better. Best results are in \textbf{bold}.}
    \label{tab:performance-language-group}
    \setlength{\tabcolsep}{7pt}
    \begin{tabular}{@{}l*{9}{c}@{}}
        \toprule
        & \multicolumn{3}{c}{High resource} & \multicolumn{3}{c}{Medium resource} & \multicolumn{3}{c}{Low resource} \\
        \cmidrule(lr){2-4}\cmidrule(lr){5-7}\cmidrule(l){8-10}
        Model & \passone & \passthree & LC & \passone & \passthree & LC & \passone & \passthree & LC \\
        \midrule
        \opusfoureight     & \textbf{51.3} & 34.3 & 90.4 & \textbf{48.1} & \textbf{33.9} & 87.2 & \textbf{44.8} & \textbf{29.6} & 86.7 \\
        \gptterrafivesix   & 45.3 & 26.3 & \textbf{98.8} & 43.9 & 25.1 & \textbf{90.9} & 36.8 & 21.9 & \textbf{89.6} \\
        \geminiprothreeone & 45.9 & 27.4 & 98.6 & 42.4 & 25.8 & 87.5 & 36.4 & 21.0 & 83.8 \\
        \qwenmaxthreeseven & 45.8 & 32.4 & 89.0 & 42.9 & 29.5 & 86.9 & 40.1 & 27.6 & 84.9 \\
        \qwenmaxthreeeight & 51.2 & \textbf{34.3} & 94.5 & 48.0 & 33.6 & 84.5 & 44.7 & 29.4 & 81.7 \\
        \bottomrule
    \end{tabular}%
\end{table}

\section{Analysis}
\label{sec:analysis}

To better understand the multilingual performance gaps, we analyze failure patterns, resource use, and language consistency.
We examine how errors vary across languages and models, how execution costs depend on language and task type, and where agents switch away from the target language.
More analysis is in \autoref{app:analysis}.

\begin{figure}[t]
  \centering
  \includegraphics[width=\linewidth]{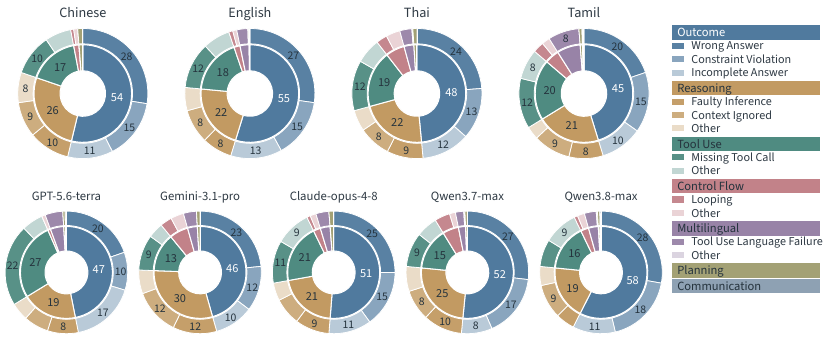}
  \caption{Primary error distributions for 1,600 failed \vitabench trajectories
  by language (top, 400 per panel) and model (bottom, 320 per panel).
  Inner rings show broad categories; outer rings show common subtypes, pooling
  rarer ones within each category. Each trajectory contributes one primary error.
  Percentages in both rings are computed over all failures per panel and rounded
  to integers. 
  See \autoref{app:analysis} for additional results.
  }
  \label{fig:analysis-taxonomy}
\end{figure}

\paragraph{Outcome errors dominate across models, but reasoning and tool use reveal distinct failure profiles.}
Outcome errors dominate all models (46--58\%; \autoref{fig:analysis-taxonomy}, bottom).
\gptterrafivesix has the largest tool-use error share (27\%); missing required calls are its most common error subtype (22\% of failures), ahead of wrong answers.
\geminiprothreeone has the largest reasoning-error share (30\%), mainly ignored context and faulty inference, while \opusfoureight has similar reasoning and tool-use error shares (about one-fifth each).
Compared with \qwenmaxthreeseven, \qwenmaxthreeeight has smaller reasoning (19\% vs.\ 25\%) and control-flow error shares (3\% vs.\ 6\%), but a larger outcome-error share (58\% vs.\ 52\%).
Explicitly language-related errors account for 3--5\% of failures across models, primarily involving language-induced tool misuse.

\paragraph{Lower-resource languages show larger shares of tool-use and control-flow failures.}
Using \geminiflashthreeseven and the hierarchical taxonomy in \autoref{app:taxonomy}, we label 1,600 failed \vitabench trajectories balanced across four languages, five models, and four domains.
Outcome errors dominate all languages, followed by reasoning and tool-use errors (\autoref{fig:analysis-taxonomy}, top).
English and Chinese have similar profiles, while Thai and Tamil show smaller outcome-error shares and larger shares of tool-use and control-flow errors, particularly unproductive loops.
Tamil has the largest explicitly language-related error share, driven by tool misuse such as arguments in the wrong language.

\begin{table}[t]
\centering
\small
\setlength{\tabcolsep}{3pt}
\caption{Resource use on \vitabench by model and language resource group, relative to English on the same tasks.
Turns count agent responses.
Constant (Const.) input counts tokens in translated fixed text: system instructions, tool definitions, and the initial user message.
Total input and output count cumulative tokens per trajectory.
Darker shading indicates greater deviation from English ($1.00\times$): red for higher usage, blue for lower.
The high-resource group excludes English.}
\label{tab:analysis-vita-cost}
\begin{tabular}{@{}l*{12}{c}@{}}
\toprule
 & \multicolumn{4}{c}{High-resource} & \multicolumn{4}{c}{Medium-resource} & \multicolumn{4}{c}{Low-resource} \\
\cmidrule(lr){2-5}\cmidrule(lr){6-9}\cmidrule(l){10-13}
Model & Turns & \shortstack{Const.\\input} & \shortstack{Total\\input} & \shortstack{Total\\output} & Turns & \shortstack{Const.\\input} & \shortstack{Total\\input} & \shortstack{Total\\output} & Turns & \shortstack{Const.\\input} & \shortstack{Total\\input} & \shortstack{Total\\output} \\
\midrule
\gptterrafivesix & \heat{0.98} & \heat{1.31} & \heat{1.20} & \heat{1.13} & \heat{0.97} & \heat{1.41} & \heat{1.26} & \heat{1.21} & \heat{0.97} & \heat{2.51} & \heat{1.92} & \heat{1.60} \\
\geminiprothreeone & \heat{1.05} & \heat{1.25} & \heat{1.23} & \heat{1.09} & \heat{1.08} & \heat{1.34} & \heat{1.41} & \heat{1.09} & \heat{1.16} & \heat{1.89} & \heat{1.90} & \heat{1.15} \\
\opusfoureight & \heat{1.03} & \heat{1.24} & \heat{1.21} & \heat{1.07} & \heat{1.03} & \heat{1.53} & \heat{1.45} & \heat{1.27} & \heat{1.05} & \heat{2.20} & \heat{2.06} & \heat{1.66} \\
\qwenmaxthreeseven & \heat{0.86} & \heat{1.15} & \heat{0.78} & \heat{0.94} & \heat{0.95} & \heat{1.23} & \heat{0.96} & \heat{1.05} & \heat{1.03} & \heat{2.10} & \heat{1.69} & \heat{1.65} \\
\qwenmaxthreeeight & \heat{0.98} & \heat{1.14} & \heat{1.10} & \heat{1.15} & \heat{0.99} & \heat{1.23} & \heat{1.20} & \heat{1.26} & \heat{1.01} & \heat{2.09} & \heat{1.80} & \heat{1.54} \\
\bottomrule
\end{tabular}

\end{table}

\paragraph{Low-resource languages incur higher token costs with a similar number of conversation turns.}
On the same \vitabench tasks (\autoref{tab:analysis-vita-cost}), every model shows its highest constant-input, total-input, and total-output ratios in the low-resource group.
Constant input rises from $1.14$--$1.31\times$ English in the high-resource group to $1.89$--$2.51\times$ in the low-resource group.
Furthermore, low-resource total input and output reach $1.69$--$2.06\times$ and $1.15$--$1.66\times$ English, respectively, while turns remain at $0.97$--$1.16\times$.
Together with the failure analysis, these results motivate more efficient and effective multilingual agents for low-resource languages.

\Needspace{18\baselineskip}
\begin{wraptable}{r}{0.48\textwidth}
\centering
\small
\setlength{\tabcolsep}{5pt}
\caption{Low-resource conversation costs on \vitabench and \textsc{DP-Shop} (\deepplanning shopping), as multiples of English usage on the same tasks.
}
\label{tab:analysis-task-cost}
\begin{tabular}{@{}l*{4}{r}@{}}
\toprule
 & \multicolumn{2}{c}{\vitabench} & \multicolumn{2}{c}{\textsc{DP-Shop}} \\
\cmidrule(lr){2-3}\cmidrule(l){4-5}
Model & Turns & Calls & Turns & Calls \\
\midrule
\gptterrafivesix & 0.97 & 0.93 & 1.14 & 1.31 \\
\geminiprothreeone & 1.16 & 1.13 & 1.71 & 1.59 \\
\opusfoureight & 1.05 & 1.03 & 1.49 & 1.66 \\
\qwenmaxthreeseven & 1.03 & 1.09 & 1.67 & 1.80 \\
\qwenmaxthreeeight & 1.01 & 1.00 & 1.68 & 1.79 \\
\bottomrule
\end{tabular}
\end{wraptable}

\paragraph{Low-resource conversation overhead varies across benchmarks and task types.}
On \vitabench, turn ratios remain at $0.97$--$1.16\times$ English and tool-call ratios at $0.93$--$1.13\times$, indicating relatively small changes in conversation length and tool-use volume.
In contrast, on \textsc{DP-Shop}, low-resource trajectories use $1.14$--$1.71\times$ as many turns and $1.31$--$1.80\times$ as many tool calls as English trajectories.
We hypothesize that this contrast partly reflects task type: \textsc{DP-Shop} requires agents to search a product catalog, where language-sensitive query matching can lead to repeated searches and query reformulation.
This search requirement may amplify low-resource conversation overhead, extending multilingual resource gaps beyond token costs to longer conversation sequences and more tool use.

\begin{wrapfigure}{r}{0.45\textwidth}
  \centering
  \captionsetup[subfigure]{font=footnotesize,skip=3pt,justification=centering}
  \begin{subfigure}{\linewidth}
    \centering
    \includegraphics[width=0.8\linewidth]{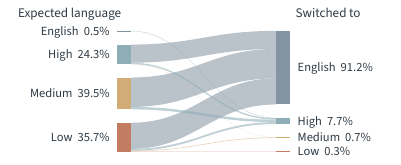}
    \caption{Switching directions across language groups.}
    \label{fig:language-switching-directions}
  \end{subfigure}
  \par\vspace{5pt}
  \begin{subfigure}{\linewidth}
    \centering
    \includegraphics[width=0.8\linewidth]{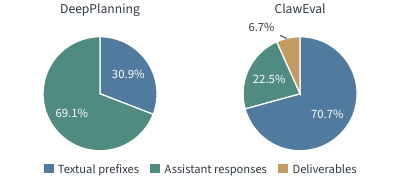}
    \caption{Switching components by benchmark.}
    \label{fig:language-switching-components}
  \end{subfigure}
    \caption{User-visible language inconsistency in 2,000 random \deepplanning and \claweval trajectories. 
    }
  \label{fig:language-consistency-switches}
\end{wrapfigure}

\paragraph{Language inconsistency is dominated by switches to English, but the affected component differs across benchmarks.}
All five models are less language-consistent on \deepplanning than on \vitabench or \claweval (\autoref{tab:performance-benchmark-family}).
We annotate switch destinations and affected components in 2,000 randomly sampled \deepplanning and \claweval trajectories flagged as language-inconsistent (\autoref{fig:language-consistency-switches}).
English accounts for 91.2\% of annotated switches; medium- and low-resource languages are rare destinations.
Inconsistency mainly occurs in textual prefixes before tool calls on \claweval and assistant responses on \deepplanning.
Manually inspecting 20 trajectories per benchmark, we find that \claweval agents commonly add English prefaces before tool calls, such as ``\textit{Let me check the weather now}''.
Similarly, 13 of the 20 inspected \deepplanning responses preface the required \texttt{<plan>}\,\dots\,\texttt{</plan>} block with English text such as ``\textit{Let me compile a travel plan now}''.
We hypothesize that these prefaces stem from training data and leave further study to future work.

\section{Conclusion}

We introduce \oursmethod, a benchmark-general agentic workflow for adapting existing agent benchmarks to new languages.
Applying \oursmethod to four complementary benchmarks yields \ours, covering a wide range of languages and domains.
We validate the effectiveness of \oursmethod and evaluate five frontier LLM agents on \ours and find that no model dominates across benchmark families and languages.
For lower-resource languages, failures shift from answer quality toward tool-use and control-flow errors, while token consumption increases substantially even on successful tasks.
We also observe that language inconsistency is most visible in textual prefixes preceding tool calls.
These findings highlight the need to evaluate task completion alongside reliability and efficiency.
\ours provides a broad, controlled testbed for advancing agents that serve users reliably and efficiently in their own languages.

\section*{AI Use Statement}

In preparing this manuscript, we used AI tools (CodeX) only to polish the language and improve the clarity of the presentation. Their use was limited to correcting grammar, rephrasing author-written sentences, and improving readability and consistency of wording. The research questions, core ideas, methodology, experimental design, analyses, and scientific conclusions of this work were developed by the human authors, and no claims, results, or references were generated by these tools. Every AI-assisted edit was reviewed for correctness by at least two human authors. We take full responsibility for the final content of this work.

\bibliography{iclr2026_conference}
\bibliographystyle{iclr2027_conference}

\appendix

\section{Language Consistency Check}
\label{app:language-consistency}

We use \geminiflashthreeseven to assess all user-visible text across the complete trajectory $\tau$ against the target language $\ell_{t}$.
The check covers assistant responses, user-visible textual prefixes preceding tool calls, and natural-language content in deliverables.
Deliverable text is included even when delivered through tools; internal reasoning, user turns, tool-call syntax, and other tool-result content are excluded.
Language consistency is evaluated independently of task completion, and a single inconsistent passage yields $\mathrm{LC}(\tau,\ell_{t})=0$ even if later responses return to $\ell_{t}$.
A trajectory can therefore be language-consistent but fail the task, or language-inconsistent but succeed; its task reward is unaffected by the LC score. The prompt used for checking language consistency is presented in \autoref{fig:language-consistency-prompt}.

\begin{figure}[p]
\centering
\fbox{%
\begin{minipage}{0.94\linewidth}
\footnotesize
\ttfamily
\raggedright
\setlength{\parindent}{0pt}
\setlength{\parskip}{0.5\baselineskip}
You are auditing whether an AI assistant replied in the language it was supposed to use.

Target language: \{language\}

Below is everything the assistant said during one task, with its tool calls and the user's turns removed. Judge ONLY the language of the assistant's own prose.

Treat these as CORRECT, never as violations: \\
- Identifiers echoed from the environment: reservation/order/user IDs, flight, train and product codes, SKUs, phone numbers, addresses, dates, times, prices and currency symbols. \\
- Tool names, function names, argument names, JSON keys, code and file paths. \\
- Structural scaffolding the output format requires. deepplanning's travel plans, for example, are emitted inside \textless{}plan\textgreater{} tags with English field labels ("Day 1:", "Current City:", "Accommodation:", "Transportation:") because the scorer parses them; only the values around those labels should be in the target language. Markdown headings, table headers and bullet markers are likewise formatting, not prose. \\
- Proper nouns that the benchmark deliberately keeps in their original form: brand names, product model names, hotel/restaurant/attraction names, airline and city names. These benchmarks preserve such names on purpose so that tool calls still match, so their presence is expected and correct. \\
- Standard loanwords and technical terms that a fluent speaker of the target language would normally leave untranslated.

Count as a VIOLATION only prose the assistant composed itself -- sentences, explanations, questions, apologies, summaries -- written in a language other than the target. An empty or near-empty sample is not a violation.

Answer with a single JSON object and nothing else: \\
\{"consistent": true|false, "share\_in\_target": \textless{}0.0-1.0\textgreater{}, "observed\_language": "\textless{}dominant language of the assistant's prose\textgreater{}", "evidence": "\textless{}at most 200 characters quoting the strongest off-language prose, or empty\textgreater{}"\}

share\_in\_target is your estimate of how much of the assistant's own prose was in the target language. Set consistent=false when meaningful prose is in the wrong language, not merely because identifiers or proper nouns appear.

--- ASSISTANT OUTPUT --- \\
\{sample\} \\
--- END ---
\end{minipage}%
}
\caption{The prompt used for trajectory-level language consistency evaluation with \geminiflashthreeseven. \texttt{\{language\}} is the target language name and \texttt{\{sample\}} is the concatenation of the assistant's own messages for one trajectory, with tool calls, tool results and user turns removed. The judge returns a JSON object; the binary \texttt{consistent} field is the label used for scoring, and \texttt{share\_in\_target} is recorded for auditing but does not enter the metric.}
\label{fig:language-consistency-prompt}
\end{figure}

\section{Quality Assurance}
\label{app:annotation}

We conduct a text-level translation audit to assess whether the multilingual variants preserve the source meaning and remain clear and natural in the target language.
For each of the 23 languages, we randomly sample 200 translatable texts, yielding 4,600 annotation records in total.
For the seven high-resource languages, each record is annotated by two human language experts, while \gptsolfivesix assesses records in the eight medium-resource and eight low-resource languages.
Each annotation assigns an integer score from 1 to 5 according to the rubric in \autoref{tab:translation-quality-rubric}.

\begin{table}[t]
    \centering
    \small
    \renewcommand{\arraystretch}{1.15}
    \caption{Five-point translation quality rubric for the text-level audit. The same criteria apply to human experts and the LLM judge.}
    \label{tab:translation-quality-rubric}
    \begin{tabular}{@{}p{0.07\linewidth}p{0.15\linewidth}p{0.69\linewidth}@{}}
        \toprule
        \textbf{Score} & \textbf{Level} & \textbf{Criteria} \\
        \midrule
        5 & Excellent & Fully faithful, complete, fluent, and natural. All relevant terminology and protected content are handled correctly, with no identifiable translation issues. \\
        4 & Good & Faithful and complete, with only minor grammatical or stylistic imperfections. These do not affect meaning, clarity of requirements, or protected content. \\
        3 & Fair & The main intent remains clear, but localized, noncritical inaccuracies or noticeably awkward wording require revision. Task-critical requirements and protected content remain intact. \\
        2 & Poor & Substantial mistranslations, omissions, or ambiguity distort the intended meaning, or an error changes a task-critical requirement or protected value. Some source meaning is still recoverable. \\
        1 & Unusable & The translation is largely incorrect, incomprehensible, or missing, or the content requiring translation is left in the wrong language. The intended meaning cannot be reliably recovered from the target-language text. \\
        \bottomrule
    \end{tabular}
\end{table}

The mean scores are 4.28, 4.02, and 3.91 for high-, medium-, and low-resource languages, respectively. 
The inter-annotator agreement (IAA) between the human experts, measured by Cohen's $\kappa$, is 0.71.
These results demonstrate the high quality of the multilingual translations in \ours.

\section{Error taxonomy and annotation}
\label{app:taxonomy}

To characterize how agents fail across languages and models, we annotate
failed \vitabench trajectories using a hierarchical error taxonomy. The
taxonomy captures both general agent errors and failures tied to
multilingual interaction. This section presents the sampling procedure,
annotation rules, complete taxonomy, and fixed prompt used to produce the
error distributions in \autoref{fig:analysis-taxonomy}.

\paragraph{Sampling and labeling}
We analyze 1,600 failed \vitabench trajectories, sampling 20 trajectories for each
combination of five models, four languages (English, Chinese, Thai, and
Tamil), and four domains. Each trajectory has task reward 0, although it
may still receive partial rubric credit. \geminiflashthreeseven annotates the
complete interaction trajectory with the model's hidden thinking removed; the
input retains user messages, assistant responses, tool calls, tool
results, and rubric verdicts.

\paragraph{Annotation prompt}
The system prompt includes the complete error taxonomy in
\autoref{tab:analysis-taxonomy}, covering eight broad categories and
39 leaf subtypes. Annotation is multi-label: every supported error is
assigned a taxonomy id together with a short quote or turn reference,
and exactly one assigned label is selected as the \emph{primary} error
most directly responsible for failure. Leaf ids are used whenever
possible, with a top-level category used only when no leaf fits. The
main-text error distributions use only the primary label, so each
trajectory contributes once.

The prompt also separates agent errors from benchmark and evaluation
issues. In non-Chinese tasks, Chinese text generated by the agent can be
labeled as a multilingual failure, while untranslated Chinese surfaced
by tools or the environment is recorded separately as a benchmark
artifact. Suspected evaluator mistakes are recorded separately, and a
novel pattern is noted when the dominant failure does not fit the
taxonomy. The complete fixed prompt is shown in
\autoref{fig:annotation-system-prompt}; at runtime,
\texttt{\{TAXONOMY\}} is replaced by the taxonomy in
\autoref{tab:analysis-taxonomy}.

\begingroup\small
\setlength{\tabcolsep}{4pt}
\renewcommand{\arraystretch}{1.05}
\begin{longtable}{@{}>{\raggedright\arraybackslash}p{0.32\linewidth}>{\raggedright\arraybackslash}p{\dimexpr0.68\linewidth-2\tabcolsep\relax}@{}}
\caption{Complete error taxonomy used in the failure analysis. Bold rows
define broad categories and their scope: Single-turn covers one response and
its input, Multi-turn covers a trajectory, and Both covers either. Subtypes
inherit the scope of their parent category.}
\label{tab:analysis-taxonomy}\\
\toprule Category / subtype & Definition \\\midrule
\endfirsthead
\multicolumn{2}{l}{\small Table \thetable\ continued}\\
\toprule Category / subtype & Definition \\\midrule
\endhead
\midrule\multicolumn{2}{r}{\small Continued on next page}\\
\endfoot
\bottomrule\endlastfoot
\textbf{Outcome}\newline\textit{Single-turn} & \textbf{The final answer is wrong, incomplete, or violates the task contract.}\\*
Wrong Answer & Final answer does not match the reference / success criterion.\\
Incomplete Answer & Answer omits required parts of the expected response.\\
Format Violation & Output does not follow the requested format or structure.\\
Constraint Violation & Answer breaks an explicit task constraint (length, scope, rules).\\
Unfaithful Answer & Answer is not supported by the provided input or context.\\
\midrule
\textbf{Reasoning}\newline\textit{Single-turn} & \textbf{The reasoning that produced the answer is flawed.}\\*
Factual Hallucination & States a fact that is false or unsupported.\\
Logical Inconsistency & Reasoning contradicts itself.\\
Faulty Inference & Draws a conclusion that does not follow from the premises.\\
Calculation Error & Makes an arithmetic or computational mistake.\\
Unfounded Assumption & Relies on an assumption with no support.\\
Context Ignored & Fails to use relevant information present in the input.\\
\midrule
\textbf{Planning}\newline\textit{Multi-turn} & \textbf{The plan or decomposition of the task is poor.}\\*
Missing Plan & Acts without forming a plan when one was needed.\\
Poor Decomposition & Breaks the task into ineffective or wrong subtasks.\\
Goal Drift & Drifts away from the original goal over the trajectory.\\
Scope Creep & Expands the task beyond what was asked.\\
Inefficient Path & Takes a needlessly long or wasteful route to the goal.\\
\midrule
\textbf{Tool Use}\newline\textit{Both} & \textbf{Errors in selecting, calling, or interpreting tools.}\\*
Wrong Tool Selected & Chooses a tool inappropriate for the subtask.\\
Hallucinated Tool & Calls a tool that does not exist.\\
Hallucinated Arguments & Invents argument values not grounded in context.\\
Malformed Arguments & Tool-call arguments violate the declared schema.\\
Missing Tool Call & Fails to call a tool that the task required.\\
Tool Output Ignored & Ignores a tool result it should have used.\\
Tool Result Misinterpreted & Misreads or misuses a tool result.\\
\midrule
\textbf{Control Flow}\newline\textit{Multi-turn} & \textbf{Errors in how the trajectory progresses or terminates.}\\*
Looping & Repeats the same step(s) without progress.\\
Oscillation & Alternates between states without converging.\\
Premature Stop & Stops before the task is complete.\\
Non Termination & Continues past the point of completion.\\
Stuck No Progress & Makes no measurable progress over several steps.\\
\midrule
\textbf{Memory}\newline\textit{Multi-turn} & \textbf{Errors in tracking state or earlier context.}\\*
Context Forgotten & Forgets information established earlier in the trajectory.\\
Self Contradiction Over Time & Contradicts an earlier statement of its own.\\
State Tracking Error & Loses track of intermediate state.\\
\midrule
\textbf{Communication}\newline\textit{Single-turn} & \textbf{Errors in how the output is communicated.}\\*
Unclear Output & Output is confusing or ambiguous.\\
Fabricated Citation & Cites a source that is fabricated or wrong.\\
Overclaiming & Asserts more certainty or capability than warranted.\\
\midrule
\textbf{Multilingual}\newline\textit{Single-turn} & \textbf{Failures tied to operating in a non-English or multilingual context.}\\*
Wrong Language Output & Responds in a different language than the user/task language (e.g. replies in English or Chinese to an Arabic task).\\
Language Mixing & Unwanted code-switching --- mixes languages within a single response when it should stay in one.\\
Script / Encoding Corruption & Garbled characters, mojibake, or wrong script --- especially in low-resource scripts (Burmese, Khmer, Tamil, Telugu).\\
Locale Convention Error & Wrong date/number/currency/unit format, name order, honorifics, or register/politeness for the locale.\\
Tool Use Language Failure & Language-induced tool misuse --- passes a localized/non-canonical value into a tool argument that expects a canonical key/ID, or mishandles an English tool result while reasoning in the task language.\\
\end{longtable}\endgroup

\begin{figure}[p]
\centering
\fbox{%
\begin{minipage}{0.94\linewidth}
\footnotesize
\ttfamily
\raggedright
\setlength{\parindent}{0pt}
\setlength{\parskip}{0.5\baselineskip}

You are an expert annotator for a multilingual tool-using agent benchmark.

You will read ONE failed agent trajectory. The task was performed in a specific language
(which may be Chinese, English, Thai, or Tamil). The benchmark's SOURCE language is
Chinese; every non-Chinese variant is a translation of the same underlying task, so any
Chinese text appearing in a non-Chinese variant is notable and you must decide whether the
AGENT produced it or the ENVIRONMENT surfaced it.

The trajectory scored reward = 0, so at least one error label is REQUIRED.

Label it against this taxonomy. Use LEAF ids (e.g. \texttt{tool\_use.tool\_output\_ignored}); use a
top-level id ONLY when the error is real but no leaf fits.

\{TAXONOMY\}

Rules: \\
1. MULTI-LABEL. Assign every label the evidence supports, typically 1--4. Do not pad. \\
2. Every label needs \texttt{evidence}: a short verbatim quote or a turn reference from the
trajectory. No evidence -> do not assign the label. \\
3. Pick exactly one \texttt{primary} label: the error that most directly caused the failure. It
must also appear in \texttt{labels}. \\
4. Judge the JUDGE too. The rubric verdicts are given; if you believe the judge was wrong
(the agent actually satisfied the requirement), say so in \texttt{judge\_disputed} and explain. \\
5. \texttt{multilingual.*} labels: assign ONLY for genuine language-induced problems. \\
\hspace*{1em}- The agent replying in the wrong language -> \texttt{wrong\_language\_output}. \\
\hspace*{1em}- The agent passing a Chinese (source-language) value into a tool argument in a
non-Chinese task -> \texttt{tool\_use\_language\_failure}. Also set
\texttt{agent\_generated\_source\_lang} to true if that Chinese string never appeared in any
earlier tool result or user turn (i.e. the agent invented it), false if it was copied from
something visible. \\
\hspace*{1em}- Untranslated Chinese appearing in TOOL RESULTS is a benchmark localisation
artifact, not an agent error: record it in \texttt{benchmark\_artifact}, do NOT label the agent
for it. \\
6. \texttt{novel\_pattern}: if the dominant failure mode is real but the taxonomy has no good
slot, describe it in one sentence. Otherwise null. \\
7. Output STRICT JSON only. No markdown fence, no commentary.

Schema: \\
\{"labels":[\{"id":"\textless{}taxonomy id\textgreater{}","evidence":"\textless{}quote/turn ref\textgreater{}","confidence":0.0-1.0\}], \\
\hspace*{1em}"primary":"\textless{}taxonomy id\textgreater{}", \\
\hspace*{1em}"agent\_generated\_source\_lang": true|false|null, \\
\hspace*{1em}"benchmark\_artifact": "\textless{}one sentence or null\textgreater{}", \\
\hspace*{1em}"judge\_disputed": "\textless{}one sentence or null\textgreater{}", \\
\hspace*{1em}"novel\_pattern": "\textless{}one sentence or null\textgreater{}", \\
\hspace*{1em}"summary":"\textless{}one sentence, in English, on why it failed\textgreater{}"\}
\end{minipage}%
}
\caption{The system prompt used for failure annotation with \geminiflashthreeseven. The
\texttt{\{TAXONOMY\}} placeholder is replaced at runtime by the complete taxonomy in
\autoref{tab:analysis-taxonomy}. The annotator receives one failed trajectory with the
model's hidden thinking removed and returns multi-label error annotations together with one
primary failure label.}
\label{fig:annotation-system-prompt}
\end{figure}


\clearpage
\section{More Analysis on Failure Types and Resource Costs}
\label{app:analysis}
\subsection{Failure Type Breakdowns}
\begin{figure}[!htbp]
\centering
\includegraphics[width=\linewidth]{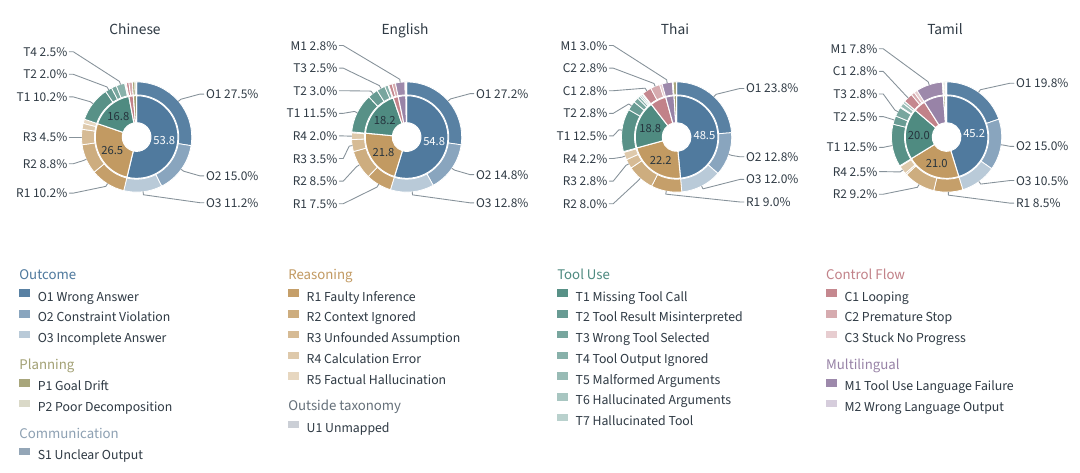}
\caption{Detailed error subtypes by language. Inner rings show broad
categories and their percentages among sampled failures; outer rings
show subtypes identified by the legend. Each language has 400 trajectories,
and each trajectory contributes one primary label.}
\label{fig:analysis-taxonomy-detail-language}
\medskip
\includegraphics[width=\linewidth]{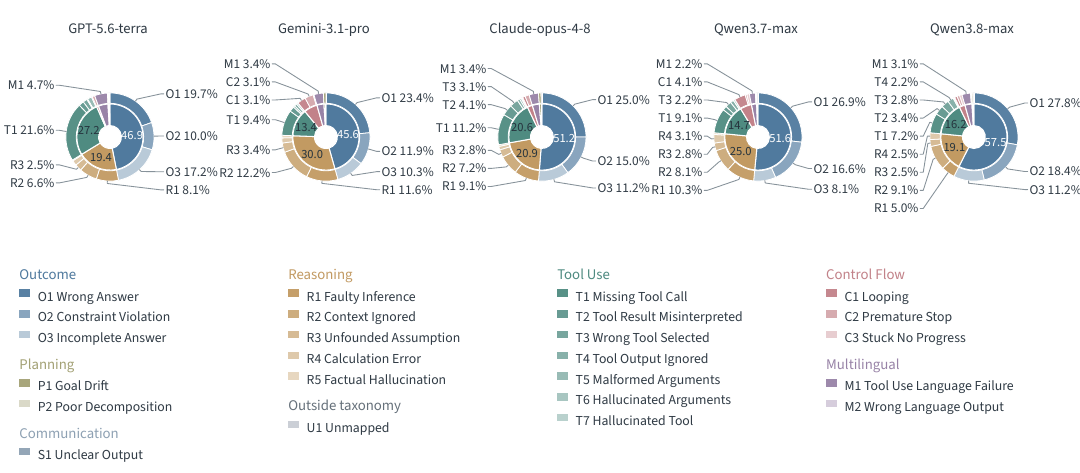}
\caption{Detailed error subtypes by model. Inner rings show broad
categories and their percentages among sampled failures; outer rings
show subtypes identified by the legend. Each model has 320 trajectories,
and each trajectory contributes one primary label.}
\label{fig:analysis-taxonomy-detail-model}
\end{figure}

\autoref{fig:analysis-taxonomy-detail-language} and~\autoref{fig:analysis-taxonomy-detail-model} provide the detailed error distributions by language and model, respectively, for the same 1,600 failed \vitabench trajectories summarized in
\autoref{fig:analysis-taxonomy}, including subtypes pooled in the main figure. Consistent with the main-text analysis, outcome errors account for the largest share across languages and
models, while Thai and Tamil have larger shares of tool use and control flow errors than English and Chinese. Tamil also has the largest share of
multilingual errors, primarily tool use language failure.
The model breakdown highlights distinct failure profiles: missing tool call is the most frequent subtype for \gptterrafivesix, while \geminiprothreeone has the largest reasoning share, mainly involving context ignored and faulty inference.
\opusfoureight has comparable shares of reasoning and tool use errors. Compared with \qwenmaxthreeseven, \qwenmaxthreeeight has smaller shares of reasoning and control flow errors but a larger outcome share.

\clearpage

\subsection{Resource accounting and paired comparisons}
\label{app:metrics}
\label{app:resource-costs}

The resource analysis in \autoref{sec:analysis} examines whether solving
the same tasks in different languages also changes execution costs. On
\vitabench, low-resource languages consume substantially more tokens than
English despite similar numbers of turns
(\autoref{tab:analysis-vita-cost}), whereas \textsc{DP-Shop} also shows
more turns and tool calls (\autoref{tab:analysis-task-cost}). These findings
motivate tracking token consumption alongside interaction counts and
comparing matched tasks across languages. This subsection details the
resource definitions, trajectory selection, and aggregation underlying
the main-text comparisons, followed by an additional analysis that holds
task success and interaction counts fixed.

\paragraph{Resource measures and trajectory selection}
Turns count generated agent responses, including responses that invoke
tools, and calls count individual tool invocations. Input and output
tokens sum provider-reported usage over agent-model calls, excluding
user-simulator and evaluator calls. The \emph{constant input} reported
in the main text is operationalized as the input token count of the
first agent-model call. It therefore captures the context present before
the first agent response, including system instructions, tool
descriptions and parameter schemas, the opening user message, and any
supplied conversation history.
For \vitabench, we retain trajectories with complete token records and tasks
with all three repetitions available in both English and the target
language. Within each model, trajectories are matched by domain, task identity,
and repetition, and repeated trajectories are averaged within each task. As in
the main-text resource analysis, the comparison includes both successful
and failed trajectories.


\paragraph{Token overhead persists on successful trajectories with equal turns and tool calls.}
We further compare paired \taubench trajectories that succeed in both
English and other languages under matching configurations and use exactly the same numbers
of turns and tool calls. \autoref{tab:analysis-exact-counts} shows that
input and output token use remain higher than English for every model in
the represented low-resource languages. 

\begin{table}[!htbp]
\centering\small
\setlength{\tabcolsep}{4pt}
\caption{Token use relative to paired English trajectories for successful
$\tau^{3}$-Bench trajectories with exactly equal turns and tool calls.
Tasks counts distinct domain--task identities; trajectory pairs counts
matched English/target repetitions.}
\label{tab:analysis-exact-counts}
\begin{tabular*}{\linewidth}{@{\extracolsep{\fill}}lrrrr@{}}
\toprule
Model & Tasks & Trajectory pairs & Input ($\times$) & Output ($\times$) \\
\midrule
GPT-5.6-Terra & 80 & 238 & 2.02 & 1.50 \\
Gemini-3.1-Pro & 103 & 442 & 1.61 & 1.38 \\
Claude-Opus-4.8 & 18 & 45 & 3.20 & 2.59 \\
Qwen-3.7-Max & 108 & 413 & 1.98 & 2.14 \\
Qwen-3.8-Max & 84 & 235 & 2.09 & 1.83 \\
\bottomrule
\end{tabular*}
\end{table}

\subsection{Repeated retrieval increases interaction costs in low-resource languages}
\label{app:shopping}

We examine whether repeated retrieval contributes to the higher
interaction costs observed in the shopping domain of \deepplanning
(\textsc{DP-Shop}; \autoref{sec:analysis}). Whereas \vitabench combines
dialogue with a simulated user and service actions such as ordering and
booking, \textsc{DP-Shop} provides the full multi-item objective at the
outset. The agent must retrieve products, inspect their attributes and
delivery constraints, and assemble a valid cart. Search ranks localized
catalog records by lexical overlap with the query, making retrieval
sensitive to query wording and script. Unsuccessful queries can therefore
lead to repeated searches and reformulation.

\paragraph{Repeated searches account for much of the additional tool use.}
\autoref{tab:shopping-retrieval} compares retrieval activity on the same
tasks in English and low-resource languages. All five models make more product searches in
low-resource languages. Across models, empty results account for
53.7\%--68.7\% of these searches, compared with 6.1\%--17.1\% in English.
Additional product searches account for 49.8\%--79.9\% of the increase
in tool calls relative to English.

The benchmark also prompts agents to verify and, if needed, correct
their carts. This stage can extend a trajectory, but 69.3\%--94.0\% of
the additional low-resource calls occur in the initial stage, before
the verification prompt. Most of the interaction overhead thus arises
during the initial attempt to assemble the cart.

\begin{table}[!htbp]
\centering\small\setlength{\tabcolsep}{3pt}
\begin{tabular*}{\linewidth}{@{\extracolsep{\fill}}lrrrrr@{}}
\toprule
 & \multicolumn{2}{c}{Searches / run} & \multicolumn{2}{c}{Empty results (\%)} & Search share \\
\cmidrule(lr){2-3}\cmidrule(lr){4-5}
Model & English & Low & English & Low & of extra calls (\%) \\
\midrule
\gptterrafivesix & 5.65 & 10.41 & 9.0 & 53.7 & 52.2 \\
\geminiprothreeone & 6.30 & 30.32 & 17.1 & 68.7 & 76.6 \\
\opusfoureight & 4.62 & 19.08 & 6.2 & 56.1 & 79.9 \\
\qwenmaxthreeseven & 4.47 & 14.22 & 6.1 & 66.8 & 49.8 \\
\qwenmaxthreeeight & 5.43 & 21.86 & 6.7 & 67.2 & 68.8 \\
\bottomrule
\end{tabular*}
\caption{Retrieval activity across all paired \textsc{DP-Shop}
trajectories, with equal weight per low-resource language (Low).
Empty-result rates are the mean number of empty product-search responses
divided by the mean number of product searches. Search share is the
increase in mean product searches divided by the increase in mean tool
calls, both relative to paired English trajectories.}
\label{tab:shopping-retrieval}
\end{table}

\paragraph{Paired successes illustrate how retrieval overhead arises.}
We manually examine paired English and low-resource trajectories that succeed on the same task.
For \qwenmaxthreeeight on task 37, the Telugu trajectory initially uses English
search phrases that return empty results. Repeated searches continue
into the cart-verification stage, and Telugu queries at calls 159--161
eventually retrieve the required candidates. Both trajectories obtain
the same four products, but Telugu uses 64 turns and 201 calls, compared
with nine turns and 25 calls in English. Product searches increase from
five in English to 158 in Telugu, of which 126 return empty results.
For \opusfoureight on task 4, the English query ``Himalayan'' immediately
retrieves the required parka. The Belarusian trajectory repeatedly
reformulates unsuccessful queries before a localized phrase retrieves
it at call 102. Both trajectories obtain the same required products and
coupon choices, but Belarusian uses 43 turns, 121 calls, and 90 searches,
including 59 empty results, compared with 13 turns, 25 calls, and four
searches in English. All 121 Belarusian calls occur before the
cart-verification prompt, illustrating that substantial retrieval
overhead can arise within the initial stage alone.
Together, the aggregate measurements and paired examples support
repeated retrieval as a source of language-related interaction overhead,
including on tasks that ultimately succeed.

\end{document}